\documentclass[journal, 10pt, letterpaper, twoside]{IEEEtran}

\usepackage[normalem]{ulem}

\usepackage{amsmath}
\usepackage{amssymb}
\usepackage{siunitx}
\usepackage[pdftex]{graphicx}
\graphicspath{{images-src/}{images-bin/}} %

\usepackage[%
sorting=none,
backend=biber, 
style=ieee,
hyperref=auto,
doi=false,
url=false,
eprint=false,
isbn=false,
maxbibnames=13,
uniquename=init
]{biblatex}

\makeatletter
\@ifpackageloaded{biblatex}{\addbibresource{bib.bib}}{\bibliography{bibliography}}
\makeatother

\usepackage[hidelinks]{hyperref}

\usepackage[nameinlink, capitalise]{cleveref}
\crefname{equation}{}{}
\Crefname{equation}{}{} 
\Crefname{figure}{Fig.}{Figs.}

\usepackage[textsize=scriptsize]{todonotes} 
\usepackage[nolist]{acronym}
\makeatletter
\AtBeginDocument
 {
   \def\ltx@label#1{\cref@label{#1}}%
   \def\label@in@display@noarg#1{\cref@old@label@in@display{#1}}%
\def\label@in@mmeasure@noarg#1{%
    \begingroup%
      \measuring@false%
      \cref@old@label@in@display{#1}%
    \endgroup}%
 } %
\makeatother

\usepackage{siunitx}

\usepackage[caption=false,font=footnotesize ,labelfont=sf,textfont=sf]{subfig}

\usepackage{booktabs}

\newbibmacro{string+doiurlisbn}[1]{%
  \iffieldundef{doi}{%
    \iffieldundef{url}{%
      \iffieldundef{isbn}{%
        \iffieldundef{issn}{%
          #1%
        }{%
          \href{http://books.google.com/books?vid=ISSN\thefield{issn}}{#1}%
        }%
      }{%
        \href{http://books.google.com/books?vid=ISBN\thefield{isbn}}{#1}%
      }%
    }{%
      \href{\thefield{url}}{#1}%
    }%
  }{%
    \href{http://dx.doi.org/\thefield{doi}}{#1}%
  }%
}

\DeclareFieldFormat{title}{\usebibmacro{string+doiurlisbn}{\mkbibemph{#1}}}
\DeclareFieldFormat[article,incollection,inproceedings]{title}%
    {\usebibmacro{string+doiurlisbn}{\mkbibquote{#1}}}

\usepackage[english]{babel}
\usepackage{csquotes}
\usepackage{blindtext}

\newcommand{\M}{\boldsymbol{M}}
\newcommand{\K}{\boldsymbol{K}}
\newcommand{\C}{\boldsymbol{C}}
\newcommand{\V}{{V}}
\newcommand{\N}{\boldsymbol{N}}

\newcommand{\q}{\boldsymbol{q}}
\renewcommand{\r}{\boldsymbol{r}}

\newcommand{\x}{\boldsymbol{x}}

\renewcommand{\vec}[1]{\boldsymbol{#1}}
\newcommand{\mtrx}[1]{\boldsymbol{#1}}

\newcommand{\bb}[1]{\mathbb{#1}}

\newcommand{\NNMS}{\acp{nnm} }
\newcommand{\NNM}{\ac{nnm} }

\definecolor{lincolngreen}{rgb}{0.11, 0.35, 0.02}

\newcommand\blfootnote[1]{%
  \begingroup
  \renewcommand\thefootnote{}\footnote{#1}%
  \addtocounter{footnote}{-1}%
  \endgroup
}

\let\FINALVERSION=1
\let\PREPRINT=1

\ifx\FINALVERSION 0
    
    \newcommand{\rev}[1]{\textbf{\color{blue}#1}}
    
    \colorlet{RevColor}{blue}
    
    \newcommand{\revrm}[1]{\rev{\textcolor{red}{\sout{#1}}}}
    \newcommand{\revrmfig}[1]{{\color{blue}#1}}
\else
    \newcommand{\rev}[1]{{#1}}

    \newcommand{\revrm}[1]{\ignorespaces}
    \newcommand{\revrmfig}[1]{}
    \colorlet{RevColor}{black}
\fi

\title{\Large \bf Locomotion of an Elastic Snake Robot via Natural Dynamics}

\author{Tristan Ehlert$^1$, Arne Sachtler$^{1,2}$, Annika Schmidt$^{1,2}$, Davide Calzolari$^{1,2}$, and Alin Albu-Schäffer$^{1,2}$%
\thanks{Manuscript received: December, 15, 2025; Revised March, 6, 2026; Accepted April, 7, 2026.}%
\thanks{This paper was recommended for publication by Editor Cosimo Della Santina upon evaluation of the Associate Editor and Reviewers' comments.}
\thanks{This work was supported by the Advanced Grant M-Runners (ID: 835284) by the European Research Council (ERC).}%
\thanks{${}^1$German Aerospace Center (DLR), Robotics and Mechatronics Center (RMC), M\"unchner Str. 20, 82234 We\ss ling, Germany}%
\thanks{${}^2$Technical University of Munich (TUM), Department of Computer
Engineering; Boltzmannstraße 3, 85748 Garching, Germany}%
}

\IEEEoverridecommandlockouts

\begin{document}
\begin{acronym}[IDA-PBC]
	\acro{bvi}[BVI]{body-velocity integral}
	\acro{bvp}[BVP]{boundary value problem}
	\acro{cad}[CAD]{computer aided design}
	\acro{com}[CoM]{center of mass}   
	\acro{cot}[CoT]{cost of transport}
	\acro{dlr}[DLR]{German Aerospace Center ({\footnotesize Ger. Deutsches Zentrum für Luft- und Raumfahrt e.V.})}
	\acro{eom}[EoM]{Equation of Motion}
	\acro{epm}[EPM]{energy per meter}
	\acro{ks}[KS]{kinematic snake}
	\acro{eks}[eKS]{elastic kinematic snake}
    \acro{nbo}[NBO]{non-brake orbit}
	\acro{nnm}[NNM]{nonlinear normal mode}
    \acro{nmm}[NMM]{natural motion manifold}
	\acro{sea}[SEA]{series elastic actuator}
\end{acronym}

\maketitle
\blfootnote{
\textbf{Copyright Note: } \copyright2026 IEEE
Personal use of this material is permitted.  Permission from IEEE must be obtained for all other uses, in any current or future media, including reprinting/republishing this material for advertising or promotional purposes, creating new collective works, for resale or redistribution to servers or lists, or reuse of any copyrighted component of this work in other works.}

\begin{abstract}
Nature suggests that exploiting the elasticities and natural dynamics of robotic systems could increase their locomotion efficiency.
Prior work on elastic snake robots supports this hypothesis, but has not fully exploited the nonlinear dynamic behavior of the systems.
Recent advances in eigenmanifold theory enable a better characterization of the natural dynamics in complex nonlinear systems.
This letter investigates if and how the nonlinear natural dynamics of a kinematic elastic snake robot can be used to design efficient gaits. 
Two types of gaits based on natural dynamics are presented and compared to a state-of-the-art approach using dynamics simulations. The results reveal that a gait generated by switching between two \aclp{nnm} does not improve the locomotion efficiency of the robot.
In contrast, gaits based on non-brake periodic trajectories (\aclp{nbo}) are perfectly efficient in the energy-conservative case.
Further simulations with friction reveal that, in a more realistic scenario, \acl{nbo} gaits achieve higher efficiency compared to the baseline gait on the rigid system.  
Overall, the investigation offers promising insights into the design of gaits based on natural dynamics, fostering further research.
\end{abstract}

\section{Introduction}
\label{sec:Introduction}
\IEEEPARstart{B}{iological} studies have shown that animals leverage their natural body dynamics and elasticities to perform periodic tasks efficiently~\cite{roberts_flexible_2011,geyer_compliant_2006}.
For example, snakes and other limbless animals use wavelike, periodic lateral motions of their bodies for locomotion \cite{gray_mechanism_1946,jayne_what_2020}.
In light of these findings, biology-inspired research has demonstrated that embedding elastic properties and exploiting the natural dynamics can also increase the efficiency of locomotion in robotic systems \cite{kashiri_overview_2018,collins_efficient_2005,alexander_three_1990,stratmann_legged_2017}.
In this letter, we intend to analyze the nonlinear natural dynamics of an elastic snake-like robot and investigate if they can be leveraged for increasing the locomotion efficiency of these periodic gaits.

\begin{figure}
  \centering
  \includegraphics[width=\linewidth]{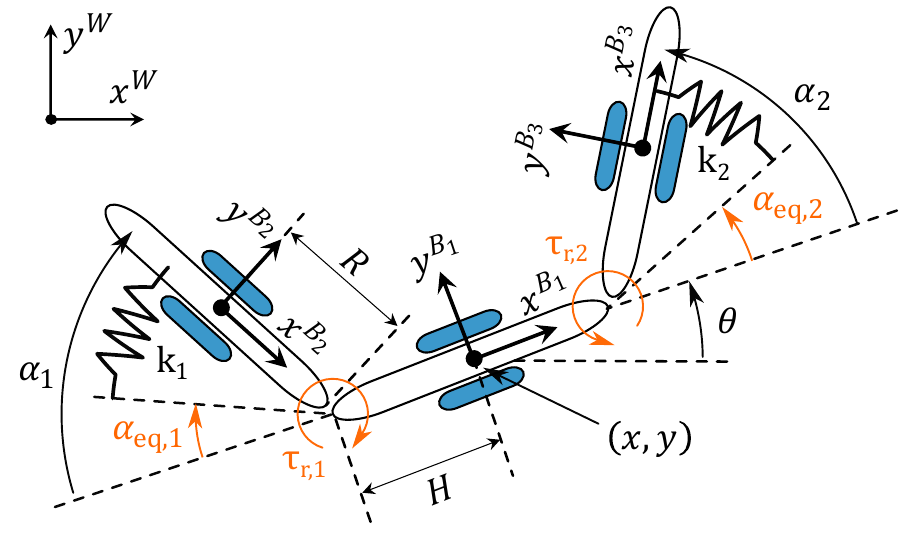}
  \caption{The model diagram of the elastic \ac{ks} with the body attached frames $ B_i $, the body lengths $ H $ and $ R $ and the configuration variables $ \q = (x,y,\theta,\alpha_1,\alpha_2) $. The inputs of the system $\vec{\alpha_{\mathbf{eq}}} = (\alpha_{\mathrm{eq,1}}, \alpha_{\mathrm{eq,2}}) $ and $\vec{\tau_r} = (\tau_{\mathrm{r,1}},\tau_{\mathrm{r,2}})$ are highlighted in orange.}
  \label{fig:KS with variables}
\end{figure}
Much work has been done investigating how the lateral undulatory gaits from snakes and other limbless animals can be transferred to robotic systems \cite{ostrowski_geometric_1998,liljeback_review_2012,shammas_geometric_2007}.

A commonly used model in this context is the {\acf{ks}} model, a three-link, planar snake model with non-holonomic constraints at each body \cite{hatton_geometric_2011,shammas_natural_2005,dear_three-link_2016}.
The geometric mechanics community has developed a range of methods for analyzing gaits for the \ac{ks} and similar kinematic systems \cite{shammas_natural_2005, hatton_geometric_2011}.
The locomotion and modelling of the \ac{ks} with a combination of actuated and passive elastic joints was investigated in \cite{dear_three-link_2016, dear_locomotive_2016, dear_locomotion_2020}, and recently \cite{raz_locomotion_2026}, but these works did not yet focus on the locomotion efficiency.
Addressing this gap, \cite{hatton_geometry_2022} presented methods for optimizing the efficiency of gaits for kinematic systems like the \ac{ks}.
A following study revealed that these methods can be adapted for actuated swimmers with passive elastic elements \cite{justus_optimal_2024}. 
These recent advances show that optimization can produce efficient gaits for locomotion in elastic snake-like systems.

Contrary to the optimization-based approaches presented so far, we analyze the natural dynamics of the system and investigate whether we can use them to design efficient gaits.
This contrasts with most previous works, since only a limited number of studies have investigated how the natural dynamics can be used to directly derive efficient gaits for elastic snake-like robots.
Only one study demonstrated that a simplified snake robot with elastic joints can be moved efficiently by exciting the eigenmodes of the linearized system \cite{ute_fast_2002}.
Meanwhile, the work in \cite{ludeke_exploiting_2020} showed that traveling wave motions exhibited in animals correspond to the natural dynamics of an idealized serial link snake model.
However, the linear modes of \cite{ute_fast_2002} have limited validity for large-amplitude oscillations, and the gaits in \cite{ludeke_exploiting_2020} are constrained to sinusoidal joint trajectories; neither work considered the full nonlinear natural dynamics.

To address this limitation, we leverage recent advances in nonlinear modal dynamics \cite{albu-schaffer_review_2020,albu-schaffer_what_2023_1} to find periodic solutions of a non-actuated lossless \ac{eks} system, and present two types of gaits that aim to exploit these natural dynamics.
The first type of gaits is based on the normal mode concept applied to nonlinear systems, exploiting the so-called \acp{nnm} \cite{albu-schaffer_review_2020}.

In contrast, the second type of gaits belong to another class of periodic orbits which we will call \acfp{nbo}.
\Acp{nbo} differ from the \acp{nnm} in that they display uninterrupted movement with non-zero kinetic energy throughout.
Such periodic orbits are exploited by passive walkers for efficient locomotion \cite{geyer_compliant_2006,collins_efficient_2005,kashiri_overview_2018}, but they have, so far, not been applied to elastic snake-like robots to enhance efficiency.

To analyze and compare the performance of the two gait types, the {\acf{eks}} shown in \cref{fig:KS with variables} is used.
This model extends the common rigid \ac{ks} model by adding elastic elements, with adjustable equilibrium positions, parallel to the joints.
Since \acp{nnm} are rigorously defined for conservative, non-hybrid systems, we adopt an ideally constrained, non-hybrid model rather than a kino-dynamic \cite{dear_three-link_2016}, or friction based formulation \cite{yona_wheeled_2019} and focus our efforts on gaits that evolve away from singularities.
The elliptical gait used on the \ac{ks} in \cite{hatton_geometric_2011} serves as the baseline against which we evaluate the natural-dynamics gaits developed here.
Performance is assessed using the mechanical \ac{cot} as a metric for locomotion efficiency.
To explore the potential real-world performance of these gaits, we also evaluate their efficiency assuming a friction model derived for the elastic kinematic snake robot ``SES-2'' in \cite{ute_fast_2002}.
The comparison of the three gait types (baseline, \NNM and \ac{nbo}) reveals drawbacks of the \NNM gaits.
Conversely, \ac{nbo}-based gaits of the \ac{eks} show promising performance compared with the baseline on the rigid \ac{ks} and the results from \cite{ute_fast_2002}. 

The structure of this letter is as follows.
The dynamic model of the \ac{eks} is derived in \cref{sec:The Elastic Kinematic Snake Model}.
In \cref{sec:Methods}, we show how the two types of gaits based on natural dynamics can be found, how the gaits are compared to each other, and how the gaits are affected by changes in the model parameters.
\cref{sec:Results} presents the results of the efficiency comparison between the two gaits developed using our new methodologies and the baseline gait from the literature.
In \cref{sec:Discussion}, the results are discussed, while a conclusion is drawn in \cref{sec:Conclusion}.

\section{The Elastic Kinematic Snake Model}
\label{sec:The Elastic Kinematic Snake Model}
The \ac{eks} model consists of three rigid bodies connected by two rotational joints, with parallel springs. 
Each body is constrained to translate exclusively along its longitudinal axis, denoted as \(x^{B_i}\), simulating the behavior of a pair of wheels attached to the midpoint of the body, provided that the lateral forces remain below the friction limit.
The diagram in \Cref{fig:KS with variables} gives an overview of the \ac{eks} with the variables and the attached coordinate systems.
The configuration $\q = \left(x,y,\theta,\alpha_1,\alpha_2 \right)$ comprises the position variables $\x = \left(x,y,\theta \right) \in SE(2)$ of body $ B_1 $, and the shape variables $ \r = (\alpha_1,\alpha_2) \in \bb{R}^2$, i.e, the angles between the central body and the other two segments.
We assume two inputs for the system: the external joint torques \(\vec{\tau_r} = (\tau_{\mathrm{r,1}},\tau_{\mathrm{r,2}})\), and an ideal position source changing the equilibria of the springs \( \vec{\alpha_{\mathbf{eq}}} = (\alpha_{\mathrm{eq,1}}, \alpha_{\mathrm{eq,2}}) \). 

The reconstruction equation is the relationship between the shape \(\r\) and the body velocities $\vec{\xi} = (\xi_x,\xi_y,\xi_{\theta})$, given in the body frame $ B_1 $.
Since the addition of the springs only influences the shape dynamics, the reconstruction equation is derived from the non-holonomic constraints of the wheels, as shown in \cite{shammas_natural_2005}, yielding
\begin{align}
\xi &= - \mathbb{A}(\r) \dot{\r} \;, \label{eq:reconstruction_eq_ks}\\
\intertext{where}
\mathbb{A}(\r) &= \frac{R}{Q} \begin{pmatrix}
  R + H \cos \left(\alpha_2\right) &
  R + H \cos \left(\alpha_1\right) \\
  0 & 0\\
  \sin \left(\alpha_2\right) &
  \sin \left(\alpha_1\right)
  \end{pmatrix} \;, \label{eq:A}\\
Q &= H \sin \left(\alpha_1-\alpha_2\right)+ R \sin \left(\alpha_1\right)- R\sin \left(\alpha_2\right). \label{eq:divisor_A}
\end{align}

The world velocities $ \dot{\x} = (\dot{x},\dot{y},\dot{\theta}) $ are calculated from the body velocities as
\begin{equation}
	\begin{pmatrix}
    \dot{x} \\
    \dot{y} \\
    \dot{\theta}
  \end{pmatrix}
  =
	\begin{pmatrix}
		\cos(\theta) & -\sin(\theta) & 0 \\
		\sin(\theta) & \cos(\theta) & 0 \\
		0 & 0 & 1
  \end{pmatrix}
  \begin{pmatrix}
    \xi_x\\
    \xi_y\\
    \xi_{\theta}
  \end{pmatrix}.
  \label{eq:worldvel_from_xi}
\end{equation}

The potential \( \V \) of the system is
\begin{equation}
\V(\r) = \frac{1}{2} (\r-\r_{\mathrm{eq}})^\top \K (\r-\r_{\mathrm{eq}}) \;,
\label{eq:potential_ks}
\end{equation}
where the stiffness matrix \( \mtrx{K} \) has the linear spring constants \( k_i \) along the diagonal.
Using the reduction method from~\cite{ostrowski_computing_1999}, the dynamics of the \ac{eks} can be expressed as
\begin{equation}
  \tilde{\M}(\r)\ddot{\r} + \tilde{\C}(\r,\dot{\r})\dot{\r} + \tilde{\N}(\r,\dot{\r}) + \frac{\partial^\top V}{\partial \r} =  \vec{\tau}_r \;,
\label{eq:shape_dynamics}
\end{equation}
where \( \tilde{\M}(\r) \) is the reduced inertia matrix, $ \tilde{\C} $ is the reduced Coriolis and centrifugal matrix, $ \tilde{\N} $ represent the conservative forces from the wheel constraints in the shape space, $ \partial^\top V/\partial \r $ are the spring forces, and $ \vec{\tau}_r $ represents the external torques applied to the joints.

The total energy of the system at state \( \left( \r,\dot{\r} \right)\) is given by
  \begin{equation}
    E(\r,\dot{\r}) = \frac{1}{2}\dot{\r}^\top\tilde{\M}(\r)\dot{\r} + \V(\r) \;.
    \label{eq:E_tot}
  \end{equation}

The parameters used in the simulations of the \ac{ks} and \ac{eks} models are given in \cref{tab:Parameters}.
We assume that each link has the same mass, and that the mass is equally distributed along the full length of each body \( l_i = 2R \; \text{or} \; 2H \).
With these parameters, the reconstruction equation and thus the model has singularities when $\alpha_1 = \alpha_2$ and $\alpha_i = \pm \pi, \quad i \in \{ 1,2 \}$.

\begin{table}
  \caption{Parameters of the \ac{ks} Model}\label{tab:Parameters}
  \centering
  \begin{tabular}{l l l}
   \toprule
   Parameter & Symbol & Value \\
   \midrule
    Link lengths & \( H = R \) & \SI{1}{\meter} \\
    Link mass & \( m_1 = m_2 = m_3 \) & \SI{1}{\kilogram} \\
    Link inertia & \( I_1 = I_2 = I_3 \) & \SI[parse-numbers=false]{\frac{1}{3}}{\kilogram\meter^2}\\
   \bottomrule
  \end{tabular}
\end{table}

\section{Methods}
\label{sec:Methods}

\begin{figure*}[h]
  \centering
  \subfloat[\label{fig:DiagGenerators}]{
    \def\svgwidth{0.30\linewidth} \small
    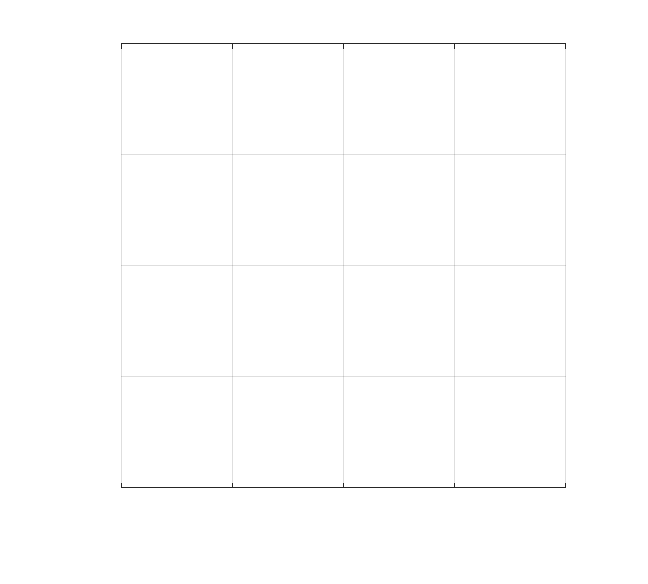}
  \subfloat[\label{fig:generator_and_modes_1}]{
    \def\svgwidth{0.3\linewidth} \small
    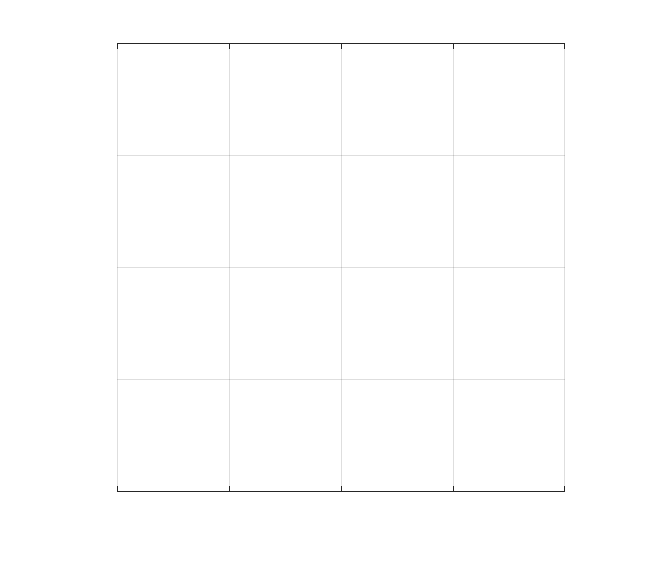}
  \subfloat[\label{fig:generator_and_modes_2}]{
    \def\svgwidth{0.30\linewidth} \small
    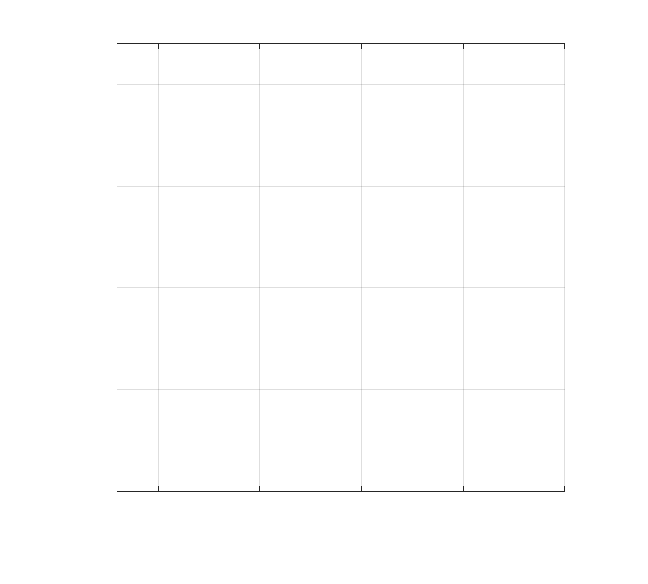}
  \caption{
    Panel (a) shows the generators from a range of equilibrium configurations of the \ac{eks} along the diagonal \(D: \alpha_1 = -\alpha_2 \).
    Panels (b) and (c) show the generators $G_\perp$ and $G_\parallel$ from two equilibrium positions (circles). The dotted lines show some exemplary nonlinear modes from these Generators. The blue dotted lines are the \acp{nnm} which the system switches between, for the gait shown in \cref{fig:modegait_example}.
  }
  \label{fig:Example_NNM_Gait}
\end{figure*}

Starting from the results of \cite{shammas_natural_2005,hatton_geometric_2011,dear_three-link_2016,yona_wheeled_2019}, we utilize the following insights regarding locomotion with the \ac{ks} as guidelines for finding and designing gaits.
\begin{itemize}
  \item Gaits shall not cross the $\alpha_1 = \alpha_2$ singularity of our model. \rev{Although singularity-crossing gaits are physically feasible on the \ac{eks}, we limit our scope to non-singularity-crossing gaits because the reconstruction equation \cref{eq:reconstruction_eq_ks} and thus our reduced model become ill-defined at this coordinate singularity \cite{dear_three-link_2016,yona_wheeled_2019}.}
  \item Gaits that are symmetric to the diagonal \( D: \alpha_1 = -\alpha_2 \) in joint space and are not self-intersecting do not produce any net rotation \cite{shammas_geometric_2007}, but can produce net displacement. 
  \item For such a symmetric, not self-intersecting gait, a larger surface area enclosed by the trajectory leads to more net translation of the robot \cite{shammas_geometric_2007,hatton_geometric_2011}. 
\end{itemize}

\subsection{Gaits based on Nonlinear Normal Modes}
\label{subsec:Gaits based on Nonlinear Normal Modes}

\Acp{nnm} and the eigenmanifold theory extend the idea of eigenmodes from linear to nonlinear systems \cite{albu-schaffer_review_2020}.
Like linear eigenmodes, \Acp{nnm} are oscillations with two turning points, where all velocities are zero.
The system moves back and forth between these turning points on a non-closed path in the configuration space.
In non-linear dynamics, these types of periodic trajectories are also referred to as brake orbits. 
As explained in detail by \cite{hatton_geometric_2011}, the net displacement of a gait can be related to the area encircled by the gait's trajectory in configuration space due to Green's form of Stokes' theorem.
Since an \ac{nnm}'s configuration trajectory does not enclose any area, moving on a \ac{nnm} does not create any net displacement,

To create \ac{nnm}-based gaits that produce net displacement, it is therefore necessary to identify modes that can be connected in sequence to form a closed trajectory.
Connecting modes requires switching points, at which the trajectories of two modes intersect. 
At these points, the equilibrium configuration \(\r_{\mathrm{eq}}\) has to be changed, such that the system can transition between modes from different equilibrium configurations.
The turning points of the modes serve as ideal switching points, as they inherently have zero joint velocities, eliminating the need for velocity matching.
In \cite{albu-schaffer_review_2020}, the collections of \ac{nnm} turning points linked to specific equilibrium configurations are called \textit{generators}.
It follows that the intersection points between two generators are shared turning points for two modes.
Thus, connectable modes can be found by checking for intersections between the generators.
One could attempt to combine several modes, as demonstrated in \cite{pollayil_planning_2022}, but we have chosen to limit our approach to combining two modes.

To find candidate gaits for the \ac{eks}, we compute eigenmanifolds and generators for a range of different equilibria and joint stiffnesses using the shooting-based method presented in \cite{albu-schaffer_review_2020}.
The results presented here are computed with a spring stiffness of \( k_1 = k_2 = \SI{10}{\newton\meter\per\radian} \).
\Cref{fig:DiagGenerators} presents generators that extend from a range of equilibrium configurations along the diagonal \( D: \alpha_1 = -\alpha_2 \).
It shows that there is always one generator $G_\parallel$ along $D$, and one that starts orthogonal to it, referred to as $G_\perp$.
The \NNMS from $G_\parallel$ are strict~\cite{albu-schaffer_strict_2021} and linear, but they are not useful for gaits.
However, the generators $G_\perp$ intersect with each other at multiple points.
The \NNMS from $G_\perp$ for equilibrium configurations along $D$ are well suited for gaits because their symmetry guarantees that the modes share both of their turning points and can thus be connected.
For equilibrium configurations off $D$, the modes and generators are only reliably symmetric to $D$ when the equilibria are.
If there are modes that share both turning points and originate from these symmetric equilibria, they could be used as gaits.
However, with our parameters, we did not find any modes from symmetric equilibria that shared both turning points and could be connected to form gaits.
By resampling the range \(\SIrange{0.21}{1.3}{\radian}\) on $D$ where there are intersections with \( 200 \) sample points, \( 1695 \) distinct gaits were found.
\Cref{fig:modegait_example} shows one of the \NNM gaits, which was found at the intersection of the two generators that are colored in \cref{fig:DiagGenerators} and shown with the modes that are connected in \cref{fig:generator_and_modes_1,fig:generator_and_modes_2}.
In simulation, the \ac{nnm} gaits are realized by initializing the system at a shared turning point, simulating until $\r(t)$ is close to the second turning point and $\dot{\r}(t) = 0$.
At that point, the spring equilibrium $\r_\textbf{eq}(t)$ is switched and the systems returns to the initial turning point along the second mode, where the equilibrium is switched again to restart the gait.

\subsection{Gaits based on \acfp{nbo}}
\label{subsec:Gaits based on Disk Orbits}
\begin{figure*}
    \subfloat[NNM gait joint trajectory \label{fig:modegait_example}]{
        \def\svgwidth{0.22\textwidth} \small
        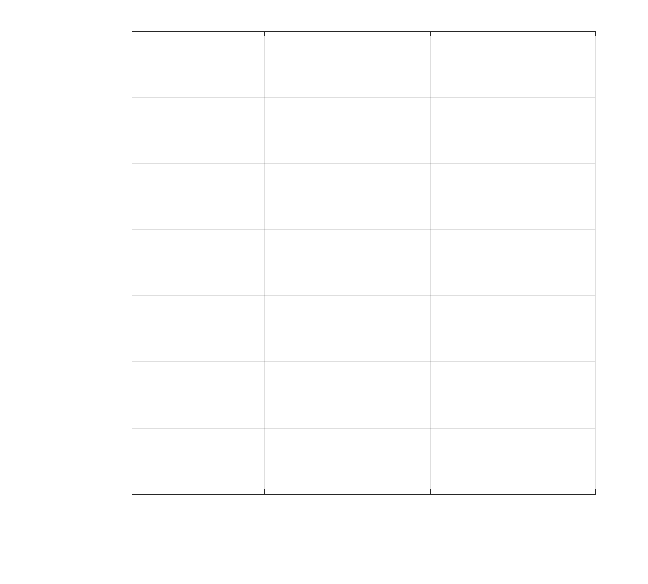}
    \subfloat[NNM gait world trajectory \label{fig:modegait_example_wt}]{
        \def\svgwidth{0.25\textwidth} \small
        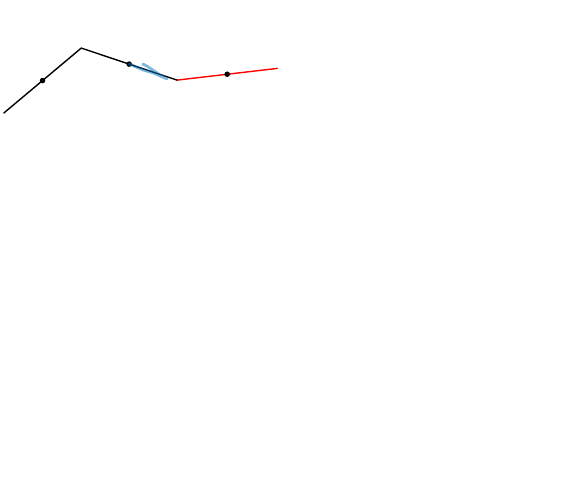}
    \subfloat[\ac{nbo} gait joint trajectory \label{fig:orbitgait_example}]{
        \def\svgwidth{0.22\textwidth} \small
        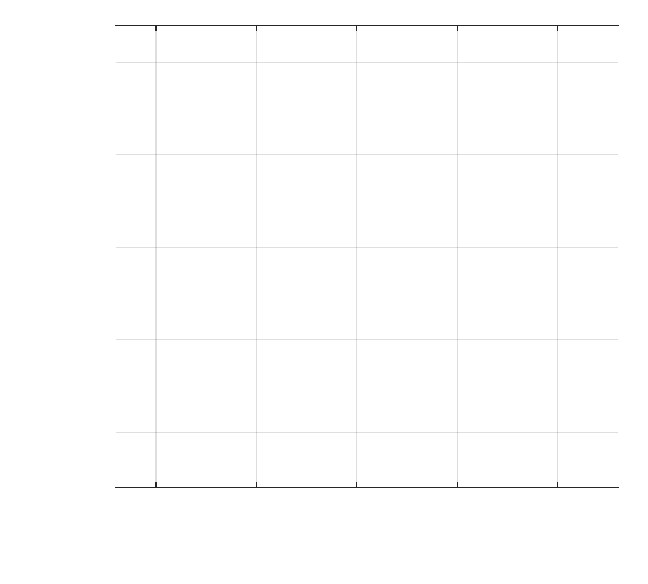}
    \subfloat[\Ac{nbo} gait world trajectory \label{fig:orbitgait_example_wt}]{
        \def\svgwidth{0.25\textwidth} \small
        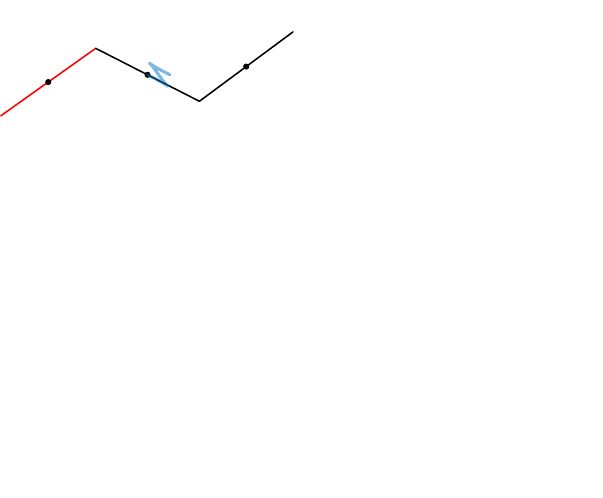}
    \caption{Examples for \NNM and \ac{nbo} gaits. On the left, panels (a) and (b) show a gait constructed by transitioning between two \acp{nnm}. At points 1 and 3, the equilibrium configurations are changed to transition between the modes. On the right, an example \ac{nbo} gait. The \ac{nbo}s start at point 1 and will continue indefinitely, assuming a conservative system. The panels (a) and (c) show the joint trajectories of the gaits. The panels (b) and (d) show the world trajectories of one period of the gaits, moving from point $1_1$ to point $1_2$, in the center and the configuration of the system at four points along the trajectory around it.}
    \label{fig:my_label}
\end{figure*}
The second type of gaits is based on periodic orbits which do not have brake or turning points (where all velocities become zero). We will call these \acfp{nbo}.
Unlike \ac{nnm}-based gaits, which require switching to achieve net displacement, \acp{nbo} enclose an area in the configuration space, meaning they produce net displacement without any external inputs.
Consequently, the \acp{nbo} can be directly used as gaits for the \ac{eks}.
As the \ac{eks}'s links are unable to rotate further than \( \pm \SI[parse-numbers = false]{\pi}{} \), all \acp{nbo} without brake points that we might find are classified as \textit{disk orbits} according to the definition in \cite{albu-schaffer_what_2023_1}. 
To find \acp{nbo}, we apply an approach similar to \cite{jahn_design_2021,albu-schaffer_what_2023_1}: using a \ac{bvp} solver, solutions \( \r(t) \) to the dynamics \cref{eq:shape_dynamics} are found to satisfy the periodicity conditions
\begin{align}
    \r(0) &= \r(T)  \;,\\
    \dot{\r}(0) &= \dot{\r}(T) \;,\\
    \vec{\tau}_r(t) &= \boldsymbol{0} \;
\end{align}
for some period time $T$.
Additionally, we require the initial value to be on the diagonal \(D: \alpha_1 = -\alpha_2 \) in joint space and the initial velocity to be normal to it
\begin{align}
    -r_1(0) &= r_2(0) \label{eq:bvp_init_pos} \;,\\
    \dot{r}_1(0) &= \dot{r}_2(0) \;. \label{eq:bvp_init_vel}
\end{align}
The constraints on the initial values \cref{eq:bvp_init_pos,eq:bvp_init_vel} are chosen such that the solutions are symmetric to $D$ and the resulting world trajectories do not produce net rotation.
The solver uses the position of the joint spring equilibria on $D$ \( -\alpha_{1,\mathrm{eq}} = \alpha_{2,\mathrm{eq}} \) and the period \( T \) as parameters.
To find \acp{nbo} that are similar to the elliptical baseline gait from \cite{hatton_geometric_2011}, its trajectory is the initial guess for the \ac{bvp} solver.
By varying the average velocity of the initial solution and the initial guess for the period, a range of different orbits can be obtained.
One problem with this approach is that it also finds \acp{nnm} that fit the constraints.
By ignoring all results that have a minimum joint velocity below \( 10^{-3} \; \SI{}{\radian\per\second} \), the \NNMS are removed.
\cref{fig:orbitgait_example,fig:orbitgait_example_wt} show one of the \acp{nbo} found with this method.

\begin{figure}[h!]
  \centering
  \def\svgwidth{\linewidth} \small
  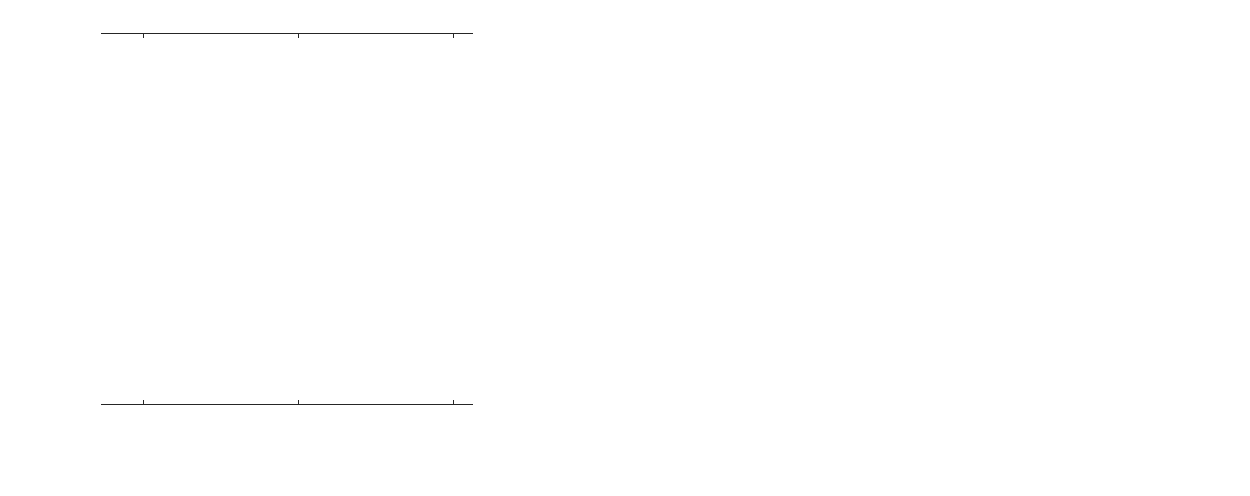    
  \caption{An example of \acp{nbo} from a \acf{nmm} containing the \ac{nbo} gait shown in \cref{fig:orbitgait_example}.}
  \label{fig:orbit_family}
\end{figure}
Once an \ac{nbo} is found, numerical continuation is applied to find similar \acp{nbo} for a desired range of energy levels \cite{raff_continuation_2025}.
In this case, the \ac{bvp} solver is employed once more, albeit without the constraints on the initial angles \cref{eq:bvp_init_pos} and velocities~\cref{eq:bvp_init_vel}.   
Instead, the original solution is utilized as an initial guess, along with a specified energy level, which is set at a fixed interval from the original solution's level. 
With this continuation, we found families of \acp{nbo} that continuously evolve with energy, as shown in \cref{fig:orbit_family}.
The families are defined in literature as \acp{nmm} of the system \cite{calzolari_exciting_2025}.
When energies become low enough, the \acp{nbo} seem to collapse into \acp{nnm}, as also seen in \cite{albu-schaffer_what_2023_1}.
Computing the \acp{nmm} helps explore the multitude of possible \acp{nbo}, and find some that are suitable as gaits for the \ac{eks}.

\subsection{Efficiency Calculations}
\label{subsec:Efficiency Calculations}
We compare the gaits using the mechanical cost of transport
\begin{equation}
    CoT = \frac{E_{\mathrm{req}}}{d \cdot m \cdot g},
    \label{eq: cot}
\end{equation}
where \(E_{\mathrm{req}}\) is energy required to move the mass of the system \(m = m_1 +m_2 + m_3\) the distance of the net displacement \(d\) at a certain average speed \( v_{\mathrm{avg}}\) in the earth's gravitation field \( g = \SI{9.81}{\meter\per\second\squared} \).

To enable a comparison of the gait efficiency with the experimental results from \cite{ute_fast_2002} we use their metric of absolute mechanical power in the system for the energy calculations.
This causes an overestimation of the actually required energy, since the energy removed from the system when braking is also counted.

As a baseline, we use the elliptical gait shown in \cite{hatton_geometric_2011} evaluated for the rigid \ac{ks}.
The energy required per period by this baseline gait is calculated by summing up the absolute mechanical work done in each joint:
\begin{equation}
  E_{\mathrm{req,B}} =  \int_{0}^{T} \sum_{i= 1}^{2} \left\lvert \tau_{r,i} \dot{r}_i \right\rvert ~ \mathrm{d}t.
  \label{eq:E_baseline}
\end{equation}

The energy required by the \NNM gaits stems from moving the equilibrium configurations of the springs.
We assume that the equilibrium position can be changed instantly, and the energy required for a switch is calculated as the difference in the potential energy stored in each spring before and after switching.
The potential difference \(\Delta V_{i} \) for each spring \( i \) is given by:
\begin{equation}
  \Delta V_{i} = \frac{1}{2} k_i \left( \left(\alpha_i - \alpha_{\mathrm{eq2},i}\right)^2 -\left(\alpha_i - \alpha_{\mathrm{eq1},i}\right)^2 \right),
  \label{eq:potential_difference}
\end{equation} 
where \( \alpha_{\mathrm{eq1},i} \) and \( \alpha_{\mathrm{eq2},i} \) are the equilibrium angles of the \( i \)th spring before and after switching.
Adding the absolute values gives the energy required for a single switch.
The energy required per period by a gait from modes is the energy put into the system to switch from mode 1 to mode 2 and back once.
Since the switching points are symmetric, the cost is equal on either side, so the total cost is:
\begin{equation}
  E_{\mathrm{req,N}} =  2 \sum_{i = 1}^{2} \left\lvert {\Delta V_{i}}\right\rvert.
  \label{eq:E_nnm}
\end{equation}

In a lossless system, the \ac{nbo} gaits, do not require any actuation once initiated and since the spring equilibrium position \(\r_{\textbf{eq}}\) is assumed to be an ideal position source, the required energy is zero
\begin{equation}
  E_{\mathrm{req,O}} =  \int_{0}^{T} \sum_{i= 1}^{2} \left\lvert \tau_{r,i} \dot{r}_i \right\rvert ~ \mathrm{d}t = 0
  \label{eq:E_orbits} \;.
\end{equation}

For a more realistic comparison, we also evaluate the \acp{nbo} and the baseline gait considering a constant rolling resistance of \( \SI{0.03}{N} \) at the bodies and joint damping of \( \vec{\tau}_d = \SI{0.023}{\newton\meter\per\radian}\dot{\r}\).
These values were found experimentally for an elastic snake robot with a similar weight to our model in \cite{ute_fast_2002}. 
The friction forces \( f_b ~\forall ~ b \in \left[1,2,3\right] \) are calculated at each body, and transformed into joint torques \( \vec{\tau}_f(\r) = \sum^3_{b=1} J_{b}(\r)^T f_b \) using  body  Jacobians \( J_b(\r) \).
The body Jacobians map the joint velocities $\dot{\r}$ to the body velocities $\xi_{b}$ of each of the 3 bodies $\xi_{b} = J_b(\r) \dot{\r}$.
For the central body this means $J_1(\r) = - \mathbb{A}(\r)$.

Without any actuation, these losses will distort the trajectories and the system would slow to a rest. 
To stay on the previously calculated trajectories, we add actuation torques $\vec{\tau}_\mathrm{comp} = \vec{\tau}_f +\vec{\tau}_d$, which compensate the losses.
These compensation torques $\vec{\tau}_\mathrm{comp}$ are added to the external torques $\vec{\tau}_r$ for calculating the required energy to move the system
\begin{equation}
    E_{\mathrm{req,D}} =  \int_{0}^{T} \sum_{i= 1}^{2} \left\lvert (\tau_{r,i} + \tau_{\mathrm{comp},i})  \dot{r}_i \right\rvert ~ \mathrm{d}t \;.
\end{equation}

\begin{figure}[ht!]
  \centering
  \def\svgwidth{0.49\textwidth} \small
  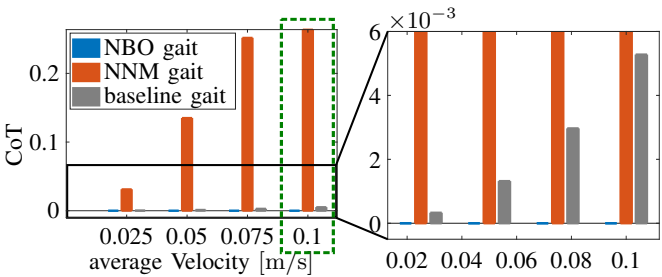
  \caption{\Ac{cot} of the \ac{nnm}, \ac{nbo} and baseline gaits, assuming no friction losses or damping. The \ac{cot} of the \ac{nbo} gaits is zero because they require no actuation when there are no losses. The trajectories of the gaits evaluated at \SI{0.1}{\meter\per\second} are shown in \cref{fig:gaits_at_0k1}.}
  \label{fig:gait_comparison}
\end{figure}
\begin{figure*}[h!]
	\centering
  \subfloat[\Ac{cot} comparison with friction.\label{fig:energy comparison with friction}]{
    \def\svgwidth{0.29\linewidth} \small
    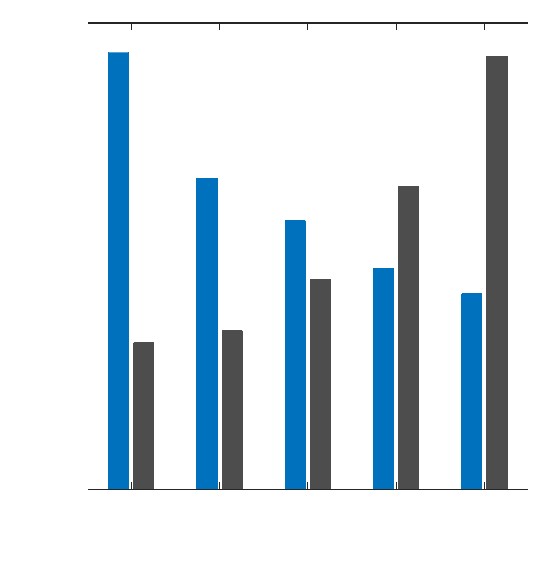
  }
  \subfloat[Gaits with \SI{0.1}{\meter\per\second} average velocity.\label{fig:gaits_at_0k1}]{ 
    \def\svgwidth{0.65\linewidth} \small
    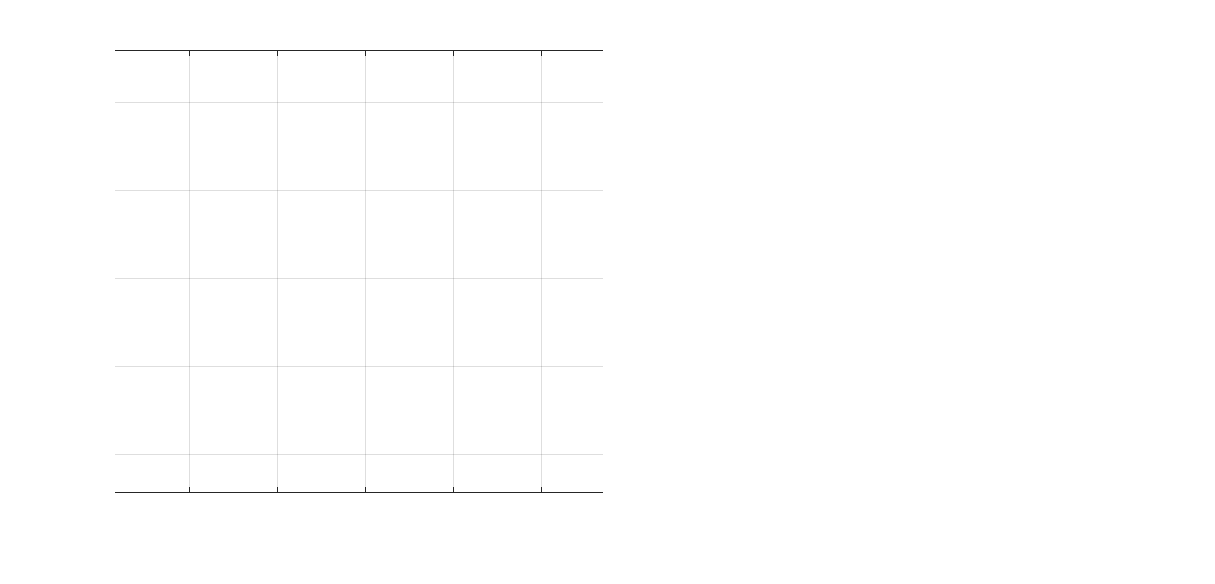
  }\\
  \subfloat[Trajectory arc length vs net displacement.\label{fig:pathlength_comp}]{
    \def\svgwidth{0.29\linewidth} \small
    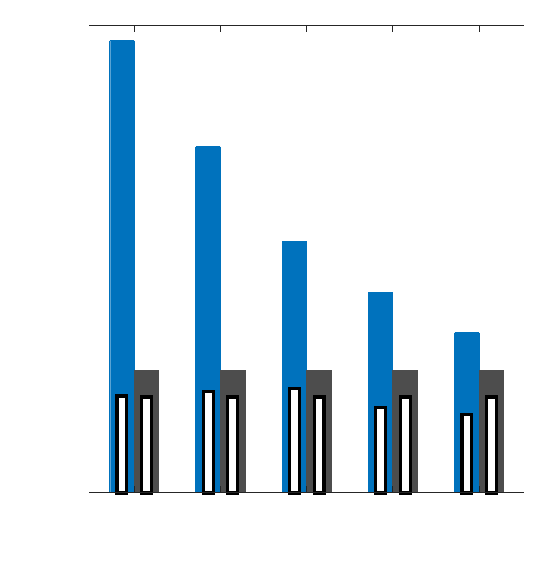
  }
  \subfloat[Gaits with \SI{0.5}{\meter\per\second} average velocity \label{fig:gaits_at_0k5}]{
    \def\svgwidth{0.65\linewidth} \small
    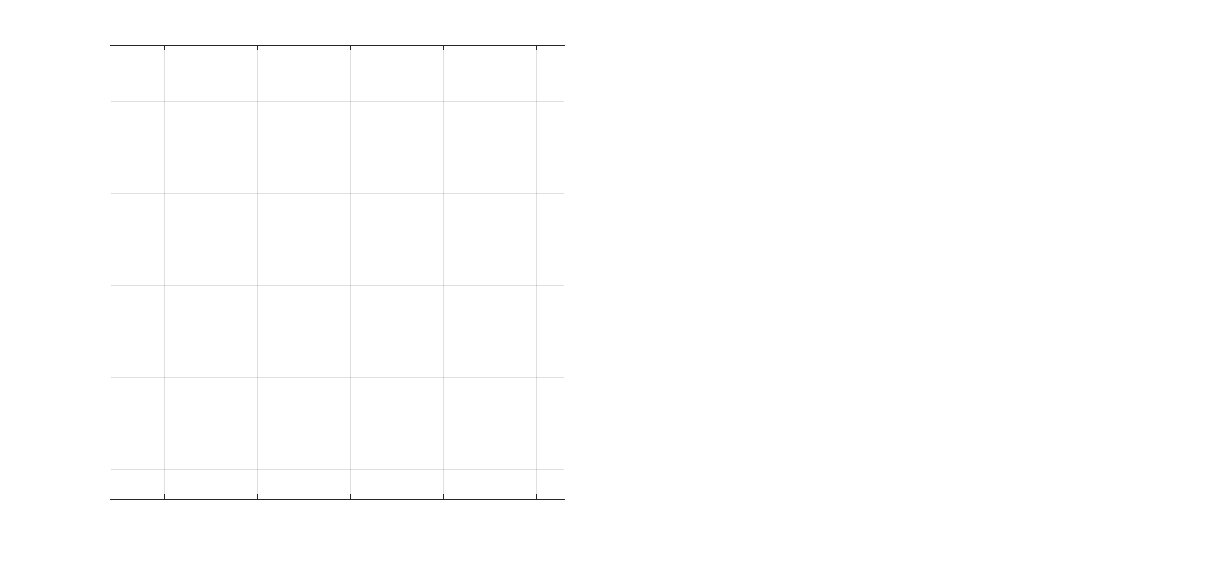
  }
  \caption{
    Panel (a): Comparing the \ac{cot} of the \acp{nbo} and the baseline with rolling resistance and joint damping as in \cite{ute_fast_2002}.
    Panel (c): Plotting the total arc length traveled by different gaits to achieve a similar net displacement (about $\SI{3}{m}$) shows that slower \ac{nbo} gaits traverse substantially longer distances than quicker \ac{nbo} gaits. The baseline gait only travels a path slightly longer than the achieved net displacement. Generally, the \ac{nbo} gaits have to travel further per net displacement than the baseline gait.
    Panels (b) and (d): World and joint trajectories of the different gait types at velocities \SI{0.1}{\meter\per\second} and \SI{0.5}{\meter\per\second}.
  }
\end{figure*}
\subsection{Parameter Influence on Gaits}
\label{subsec:parameter_influence}
Evaluating the efficiency of the natural dynamics-based gaits for only one set of parameters would result in a very limited view of the performance that is possible on the \ac{eks} by exploiting the natural dynamics.
To avoid recalculating the generators and \acp{nbo} for a whole range of parameter sets, we investigated the influence of the parameters, e.g., joint stiffness \(k\), link length \(l\), and link mass \(m\) on the \NNMS and \acp{nbo}.
We simulated ten of the \NNMS with stiffnesses between \(1\) and \SI{100}{\newton\meter\per\radian}, masses between \( 0.1 \) and \SI{2}{\kilogram} and link lengths between \( 0.05 \) and  \SI{2}{\meter} keeping the stiffnesses equal between joints, and the link-mass and link-lengths equal for each of the links.

We found that with these changes, the joint trajectories~$\r(t)$ always traverse the same path, just the period $T$ changes.
From the simulated data we empirically derived the following scaling relationships for the resulting period $T$, and the displacements per period $d$ depending on the change of the parameters $k$, $l$ and $m$:
  \begin{align}
    T_{\mathrm{new}} &= T_{\mathrm{old}} \frac{l_{\mathrm{new}}}{l_{\mathrm{old}} } \sqrt{\frac{k_{\mathrm{old}}}{k_{\mathrm{new}}}\frac{m_{\mathrm{new}}}{m_{\mathrm{old}}}}, \label{eq:T_new}\\
    d_{\mathrm{new}} &= d_{\mathrm{old}} \frac{l_{\mathrm{new}}}{l_{\mathrm{old}}}, \label{eq:d_new}
\end{align}
Combining \cref{eq:T_new} and \cref{eq:d_new} produces a relation for the average velocity \(v_{\mathrm{avg}}\) of the gaits:
\begin{align}
    v_{\mathrm{avg},new} &= v_{\mathrm{avg},old}\sqrt{\frac{k_{\mathrm{new}}}{k_{\mathrm{old}}}\frac{m_{\mathrm{old}}}{m_{\mathrm{new}}}} \label{eq:v_avg_new}\;
\end{align} 
From \cref{eq:potential_difference} and \cref{eq:E_nnm} a relation for the required energy \(E_{\mathrm{req,N}}\) can be derived:
\begin{align}
    E_{\mathrm{new}} &= E_{\mathrm{old}} \frac{{k_{\mathrm{new}}}}{k_{\mathrm{old}}} \;. \label{eq:E_new}
  \end{align}
With these relations, we are able to predict the speed and efficiency of the \NNM gaits at other parameters.
This allows for more generalized statements when evaluating the efficiency of \NNM gaits on the \ac{eks}.
Notably, \cref{eq:T_new,eq:d_new} can also be derived analytically, following \cite{landau_mechanics_1976} to estimate the $T$ between the turning points and using non-dimensionalization to extract the parameter scaling.\footnote{\rev{The full derivation is provided in the appendix of our preprint, available at \url{https://arxiv.org/abs/XXXX.XXXXX}.}}

Testing these relations for a few of the \acp{nbo} showed that they hold up there too, except now the energy scaled in \cref{eq:E_new} is the energy level calculated with \cref{eq:E_tot}. 
Like for the \acp{nnm}, this enables prediction of the speed and the energy level of gaits with changed parameters, which in turn allows for easier calculation of the \acp{nbo} at other parameters, since the period time and energy level guesses used in solving the \ac{bvp} are already correct.

\section{Results}
\label{sec:Results}
\Cref{fig:gait_comparison} shows the \ac{cot} of gaits at a range of velocities, assuming no friction losses.
For the \NNM gaits, the lowest \ac{cot} \(0.03\) occurs at the lowest velocity of \( \SI{0.025}{\meter\per\second} \).
The highest \ac{cot} is \(0.26\), occurring at the highest velocity \( \SI{0.1}{\meter\per\second} \).
The baseline's \ac{cot} increases from \(0.0003\) to \(0.005\) across the range of velocities.
The \ac{nbo} gaits have zero \ac{cot} across all velocities, as they require no actuation in a conservative system once initiated.
While zero losses are unrealistic in real systems, it indicates the potential of the \acp{nbo} to be more energy efficient than the baseline.
The \NNM gaits, however, are substantially less efficient than either and do not seem to be a preferable method for efficient locomotion with the \ac{eks}, which is why they were excluded from the following comparison with friction.

In the more realistic comparison with friction, shown in \Cref{fig:energy comparison with friction}, the performance of the \ac{nbo} gaits versus the baseline gait varies substantially with velocity.
Notably, at low velocities \( \SI{0.1}{\meter\per\second} \), the \ac{cot} for \ac{nbo} gaits was found to be about \( 3 \) times higher than that of the baseline gait, with a \ac{cot} of \( 0.05 \) for the \ac{nbo} gait compared to \( 0.14 \) for the baseline.
However, as velocity increases, the \ac{nbo} gait's \ac{cot} decreases, whereas the baseline gait's \ac{cot} steadily rises.
Specifically, at velocities above \( \SI{0.4}{\meter\per\second} \), the \ac{nbo} gaits are more efficient than the baseline, with their \ac{cot} decreasing to \( 0.06 \) at \( \SI{0.5}{\meter\per\second} \), outperforming the baseline gait's \ac{cot} of \( 0.14 \) at the same velocity.

\Cref{fig:gaits_at_0k1} shows the world trajectories of the three different gaits that are compared in \cref{fig:gait_comparison} at $\SI{0.1}{\meter\per\second}$.
The comparison shows that, at this velocity, the new gaits have substantially smaller joint trajectories and lower displacements per period $d$ compared to the baseline gait.
At higher velocities, both the size of the joint trajectory as well as the $d$ of \ac{nbo} gaits increase.
For example, the \ac{nbo} gait at $\SI{0.5}{\meter\per\second}$ shown in \cref{fig:gaits_at_0k5}, almost quadruples $d$ compared to the gait at $\SI{0.1}{\meter\per\second}$ but remains less than half compared to the baseline trajectory.

\section{Discussion}
\label{sec:Discussion}
In the simulations, we evaluated the locomotion efficiency of two dynamics-based gaits on the \ac{eks} and a baseline gait on the rigid \ac{ks}.
Our results show that the \ac{nnm}-based gaits are substantially less energy efficient than the baseline gait and the \ac{nbo} gaits.
The higher \ac{cot} of these gaits is due to the substantially lower net displacement per period compared to the baseline gait, paired with the high costs of switching the equilibrium configuration twice per period.
Considering the scaling equations \cref{eq:T_new,eq:d_new,eq:v_avg_new,eq:E_new}, we conclude that tuning the \ac{eks} parameters is unlikely to yield improved \ac{nnm}-based gaits.

In contrast, the \ac{nbo} gaits appear to be cost-free for a conservative system (\cref{fig:gait_comparison}). 
A real system, however, will not be conservative.
The simulations incorporating rolling and joint friction offer a preliminary understanding of the anticipated mechanical efficiency:
For velocities up to \(\SI{0.3}{\meter\per\second} \), the \ac{nbo} gaits are less efficient than the baseline (\cref{fig:energy comparison with friction}).
This can be attributed to the substantial disparity between the arc length of the configuration space trajectory and the resulting net displacement of the \ac{nbo} gaits at lower speeds (\cref{fig:pathlength_comp}), leading to substantially higher rolling resistance losses than for the baseline.
As the average velocity increases and the joint trajectories of the \ac{nbo} gaits expand, the ratio of net displacement to arc length increases, thereby reducing the relative amount of frictional losses.
In contrast, the rolling losses of the baseline gait remain constant across velocities, as rolling resistance is assumed to be velocity invariant and the trajectory is fixed, merely being traversed more quickly to achieve higher speeds.
However, the actuation torques required by the baseline gait to drive the joints along the desired trajectory increase with velocity.
This leads to the general increase in the baseline gait's \ac{cot} shown in \cref{fig:energy comparison with friction}.
Considering this, our results show that the larger and quicker \acp{nbo}, where the difference in rolling losses is reduced, are more efficient than the baseline gait, because the trajectories do not required further actuation, since they are solutions of the system's natural dynamics.
More generally, our results show that for efficient locomotion \acp{nbo} with a high net displacement to traversed arc length ratio should be preferred.
Notably, the observed velocity dependency of the efficiency is not due to resonance between input and system, as it was observed in elastic swimmers \cite{zigelman_dynamics_2024,justus_optimal_2024}, but is a result of the chosen friction model.

A comparison between the proposed \ac{nbo} gaits and the self-excited gait based on linearized dynamics from \cite{ute_fast_2002} highlights the possible advantage of considering the full nonlinear dynamics.
At a speed of \SI{0.5}{\meter\per\second}, the \ac{nbo} gait achieves a \ac{cot} of \(0.06\), substantially lower than the \(0.22\) reported for the SES2 robot in \cite{ute_fast_2002}.
This comparison must be interpreted with caution, as the SES2 model includes losses due to normal wheel slip, whereas our model assumes ideal no-slip constraints.
Approximating such ideal constraints in hardware requires ensuring that the necessary constraint forces are physically realizable. Despite these differences, the results demonstrate the potential benefits of using nonlinear dynamics over linearized approximations for efficient locomotion.

There are a few limitations to our study that should be highlighted and may be addressed in future work.

Firstly, the baseline gait taken from \cite{hatton_geometric_2011} was not designed or optimized for locomotion efficiency.
Using the tools from \cite{justus_optimal_2024,hatton_geometry_2022}, one could possibly find a more optimal gait for the \ac{ks} to compare the \acp{nbo} against.
More importantly, the analysis is limited to non-singularity-crossing gaits, as our model's non-hybrid dynamics become ill-defined at $\alpha_1 = \alpha_2$.
Singularity-crossing gaits yield substantially higher net displacement \cite{dear_locomotive_2016,hatton_geometry_2022}.
Our \ac{nnm} trajectories seem to align transversely with the singularity line \cref{fig:generator_and_modes_1,fig:generator_and_modes_2}, possibly allowing for their extension across the singularities \cite{dear_locomotive_2016}.
However, extending our work to singularity-crossing gaits requires a hybrid model \cite{dear_three-link_2016} or relaxed constraints \cite{yona_wheeled_2019}, which we intend to address in future work.

Secondly, we computed the friction and damping losses based on the resulting trajectories from the frictionless system, rather than simulating the dynamics that include friction and damping.
This was possible because, for our model, the friction and damping forces could be directly compensated by external joint torques without impacting the trajectories.
For other kinds of dissipation, e.g., if the no-slip constraints were relaxed to a friction model, this might not be possible, and the dissipation would influence the shape of the orbits.
Additionally, simulating the system with friction could be challenging since the rolling resistance forces are discontinuous as the bodies of the \ac{ks} move forwards and backwards.
Investigations into finding optimal gaits for dissipative systems and the influence of dissipation on the natural gaits are ongoing topics of our research group \cite{griesbauer_discovering_2025} and might be applied to the \ac{eks} in future work.

Finally, we identified two intriguing relationships between \acp{nnm} and \acp{nbo} in our results, which may warrant further theoretical investigation.
In \cref{subsec:Gaits based on Disk Orbits}, we observe that the \acp{nbo} seem to collapse into \acp{nnm}.
This behavior might be related to the transition between back-and-forth and full rotating orbits that can be seen in a pendulum, as the energy level increases \-- a similar behavior was also observed in a double pendulum
\cite{albu-schaffer_what_2023_1}.
There is also a visual similarity between the joint trajectories of the \ac{nnm} gait and the high-energy \acp{nbo}, \cref{fig:modegait_example,fig:orbit_family}, respectively.
Further investigation of these two phenomena may reveal insights into controlled transitioning between \acp{nnm} and \acp{nbo}, thereby increasing the flexibility of systems that utilize diverse natural dynamics trajectories for locomotion or other tasks.

Despite these limitations and open questions, our simulation results demonstrated that gaits based on nonlinear natural dynamics can be advantageous for fast, energy-efficient locomotion.
Consequently, the design of robots capable of achieving efficient and fast locomotion should take into account the nonlinear natural dynamics and their effect on the attainable gaits.
In the future, we aim to develop a dynamics-based co-design process for robotic systems, enabling them to leverage their nonlinear natural dynamics for more efficient locomotion.
This co-design process should involve identifying optimal and feasible parameters, such as masses, link lengths, and joint stiffness, within a desired range of velocities and other design constraints.
The selection of appropriate parameters is critical, as they determine the variety of \acp{nbo} theoretically available and are relevant to the feasibility of the \acp{nbo}.
For instance, the mass directly influences the \ac{nbo} gait velocity and the lateral force limit that must be respected under the zero-slip assumption.
The co-design process should also include optimizing the shape of the gaits using the available parameters.
Using this process, we aim to design and build a robot that enables validation of the findings in hardware.

\section{Conclusions}
\label{sec:Conclusion}
We presented two methods for using the natural dynamics of an elastic snake robot for locomotion.
Our simulations showed that \ac{nnm}-based gaits are unable to improve the efficiency of this system.
\Acp{nbo} as gaits, on the other hand, offer theoretically perfectly efficient locomotion.
Even when friction and damping losses are considered, the simulations show improvement over the baseline gait on the rigid robot at higher velocities.
The results so far support the idea that leveraging the natural dynamics of an elastic snake robot can improve locomotion efficiency.
Through a co-design process that shapes and optimizes the natural dynamics of a system for specific tasks or trajectories, we would be one step closer to designing and building robotic systems that can fully leverage their nonlinear natural dynamics for more efficient locomotion.

\rev{\section*{Acknowledgment}
We would like to thank the anonymous reviewers for their valuable feedback and suggestions.}

\newpage
\appendix
\section{Analytical Derivation of the Parameter Scaling Relations}
\label{app:scaling_derivation}

Here we show the analytical derivation of the scaling relations \cref{eq:T_new,eq:d_new}, mentioned in \cref{subsec:parameter_influence}.

For a \NNM, the trajectory between two consecutive turning points traces a fixed curve in joint space, which we parameterize by a curvilinear coordinate $s\in[0,1]$, with $\r(s)$ such that  $s=0$ and $s=1$ denote the turning points.
Considering the dynamics of $s(t)$, the total energy can be expressed as a function of the single pair $(s(t), \dot{s}(t))$:
\begin{equation}
    E_0 = \frac{1}{2}\bigl((\mathbf{r}')^\top \tilde{M}\, \mathbf{r}'\bigr)\dot{s}^2
          + \frac{1}{2}\sum_i k\bigl(\alpha_i(s) - \alpha_{\mathrm{eq},i}\bigr)^2 \;,
    \label{eq:energy_s}
\end{equation}
where $\mathbf{r}' = \mathrm{d}\mathbf{r}/\mathrm{d}s$.
Following~\cite{landau_mechanics_1976}, the period between two turning points can be expressed as
\begin{equation}
  T = \int_{0}^{1}
      \frac{\sqrt{(\mathbf{r}')^\top \tilde{M}\, \mathbf{r}'}}
           {\sqrt{2E_0 - \sum_i k\bigl(\alpha_i(s)-\alpha_{\mathrm{eq},i}\bigr)^2}}
      \, \mathrm{d}s \;.
      \label{eq:app_landau_period}
\end{equation}
Non-dimensionalizing the reduced mass matrix as 
\begin{equation}
    \tilde{M} = (\tilde{M})_\mathrm{nd}\,ml^2
\end{equation}
and substituting the turning-point condition 
 \begin{equation}
    E_0 = \tfrac{1}{2}k\sum_i\Delta\alpha_i^2
 \end{equation} 
factors all dimensional quantities out of the (parameter-invariant) trajectory integral:
\begin{equation}
  T = l\sqrt{\tfrac{m}{k}}\int_{0}^{1}
      \frac{\sqrt{(\mathbf{r}')^\top (\tilde{M})_\mathrm{nd}\,\mathbf{r}'}}
           {\sqrt{\sum_i\Delta\alpha_i^2
                  - \sum_i\bigl(\alpha_i(s)-\alpha_{\mathrm{eq},i}\bigr)^2}}
      \, \mathrm{d}s \;,
\end{equation}
which immediately yields \cref{eq:T_new}.

For the displacement scaling \cref{eq:d_new}, the same procedure is applied to the reconstruction equation \cref{eq:reconstruction_eq_ks,eq:A,eq:divisor_A}.
Non-dimensionalizing the local connection as
\begin{equation}
    \mathbb{A} = \mathbb{A}_\mathrm{nd}\,l
\end{equation}
and using the trajectory parameterization $\dot{\mathbf{r}} = \mathbf{r}'(s)\dot{s}$, the displacement per period becomes
\begin{equation}
    d = \int_0^T \xi_x \, \mathrm{d}t
      = -l \int_0^1 \bigl[\mathbb{A}_\mathrm{nd}(\mathbf{r})\,\mathbf{r}'(s)\bigr]_x \, \mathrm{d}s \;,
\end{equation}
where the integral depends only on the trajectory.
The linear dependence on $l$ immediately yields \cref{eq:d_new}.

\clearpage
\printbibliography

\end{document}